\documentclass[letterpaper, 10 pt, conference]{ieeeconf}  

\IEEEoverridecommandlockouts                              

\usepackage[nolist,nohyperlinks]{acronym}

\usepackage{todonotes}
\usepackage{graphicx}
\usepackage{caption}
\usepackage{subfig}
\usepackage{gensymb}
\usepackage{booktabs}

\usepackage{amssymb}
\usepackage{amsmath}

\newcommand{\unit}[1]{\,\mathrm{#1}}

\title{\LARGE \bf
Aerial Picking and Delivery of Magnetic Objects with MAVs
}

\author{Abel Gawel\authorrefmark{1}, Mina Kamel\authorrefmark{1}, Tonci Novkovic\authorrefmark{1},\\ Jakob Widauer, Dominik Schindler, Benjamin Pfyffer von Altishofen, Roland Siegwart, and Juan Nieto\\
\authorblockA{\authorrefmark{1}The authors contributed equally to this work.}\thanks{Authors are with the Autonomous Systems Lab, ETH Zurich. \tt\small \{abel.gawel, mina.kamel, tonci.novkovic, dominik.schindler\}@mavt.ethz.ch, \tt\small \{jwidauer, bpfyffer\}@student.ethz.ch, \tt\small \{rsiegwart, nietoj\}@ethz.ch}.}

\begin{document}

\maketitle
\thispagestyle{empty}
\pagestyle{empty}

\begin{abstract}

Autonomous delivery of goods using a \ac{MAV} is a difficult problem, as it poses high demand on the \ac{MAV}'s control, perception and manipulation capabilities. This problem is especially challenging if the exact shape, location and configuration of the objects are unknown.

In this paper, we report our findings during the development and evaluation of a fully integrated system that is energy efficient and enables \ac{MAV}s to pick up and deliver objects with partly ferrous surface of varying shapes and weights. This is achieved by using a novel combination of an electro-permanent magnetic gripper with a passively compliant structure and integration with detection, control and servo positioning algorithms. The system's ability to grasp stationary and moving objects was tested, as well as its ability to cope with different shapes of the object and external disturbances. We show that such a system can be successfully deployed in scenarios where an object with partly ferrous parts needs to be gripped and placed in a predetermined location.

\end{abstract}


\section{INTRODUCTION}
Fast and customized delivery of goods is a major trend in transportation industry. \ac{MAV}s are expected to be an important component in the future of autonomous delivery and are a means of transportation at the edge of consumer market entry \cite{zhang2014scheduling}. Most solutions for handling goods with \ac{MAV}s rely on mechanical gripping devices, as these can be realized lightweight and energy-efficient for high payloads \cite{pounds2011grasping, lindsey2011construction}.
However, mechanical grippers usually require highly precise positioning of the gripper with respect to the object to yield a safe form closure or friction fit.
High positioning accuracy cannot always be achieved with the \ac{MAV} control alone, due to environmental disturbances, making either human intervention necessary, or requiring sophisticated additional actuators.
Furthermore, the gripper design depends on the geometry of the objects to grip \cite{backus2014design} making it necessary to use standard transportation containers or facilitate a variety of different mechanical grippers to enable \ac{MAV}s to reliably grip differently shaped objects.
Ferrous objects are interesting because they can be attracted by magnets.
For gripping, these material properties can be exploited. In this case positioning accuracy can be considerably lower as a natural attraction force is generated between the magnetic gripper and ferrous material. However, using electro-magnets requires a constant power-supply to generate the magnetic field. On the other hand permanent magnets do not consume power, but they are problematic for releasing attracted objects. A new class of electro-permanent magnets overcomes both these limitations by providing a switchable permanent magnet \cite{knaian2010electropermanent}.

\begin{figure} [t!]
\centering
\subfloat[
\label{fig:teaser_a}]{\includegraphics[width=0.22\textwidth]{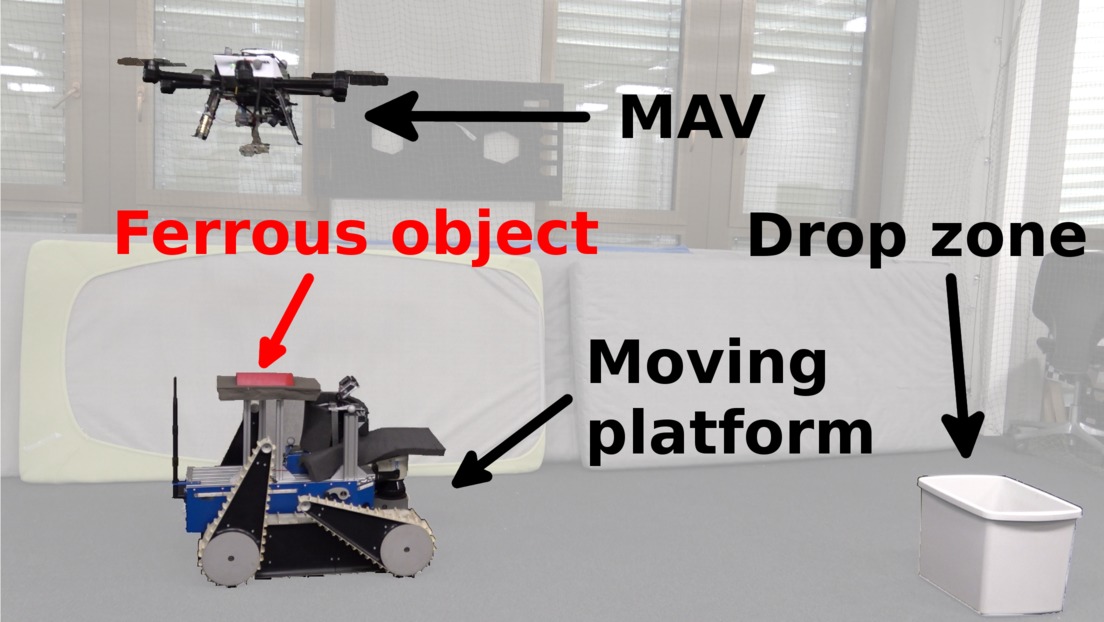}}
\subfloat[\label{fig:teaser_b}]{
\includegraphics[width=0.22\textwidth]{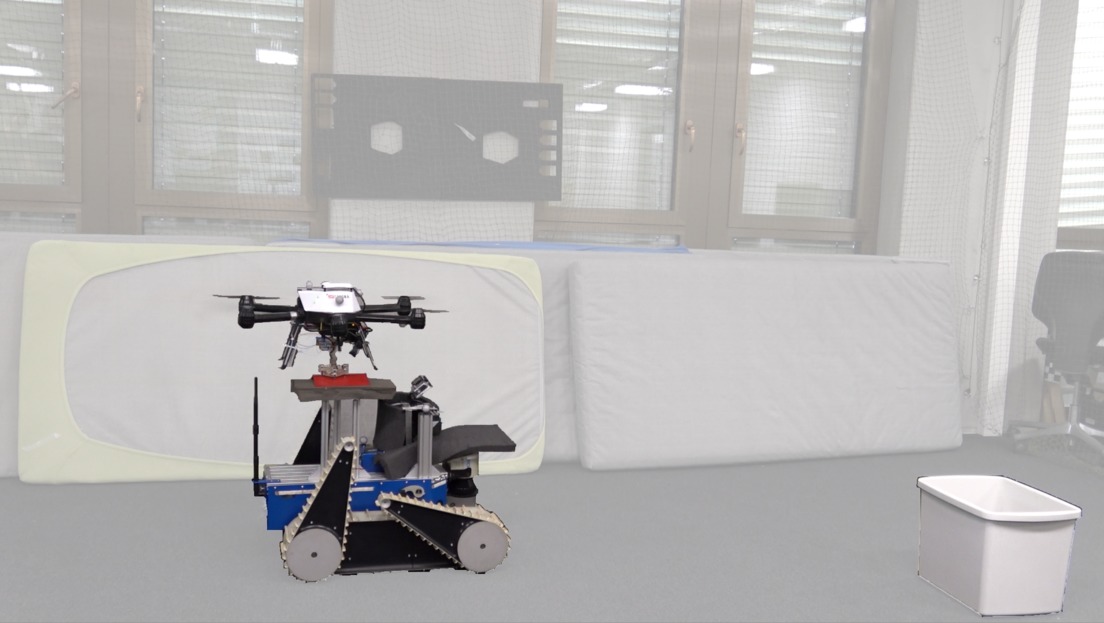}}\\
\subfloat[\label{fig:teaser_c}]{
\includegraphics[width=0.22\textwidth]{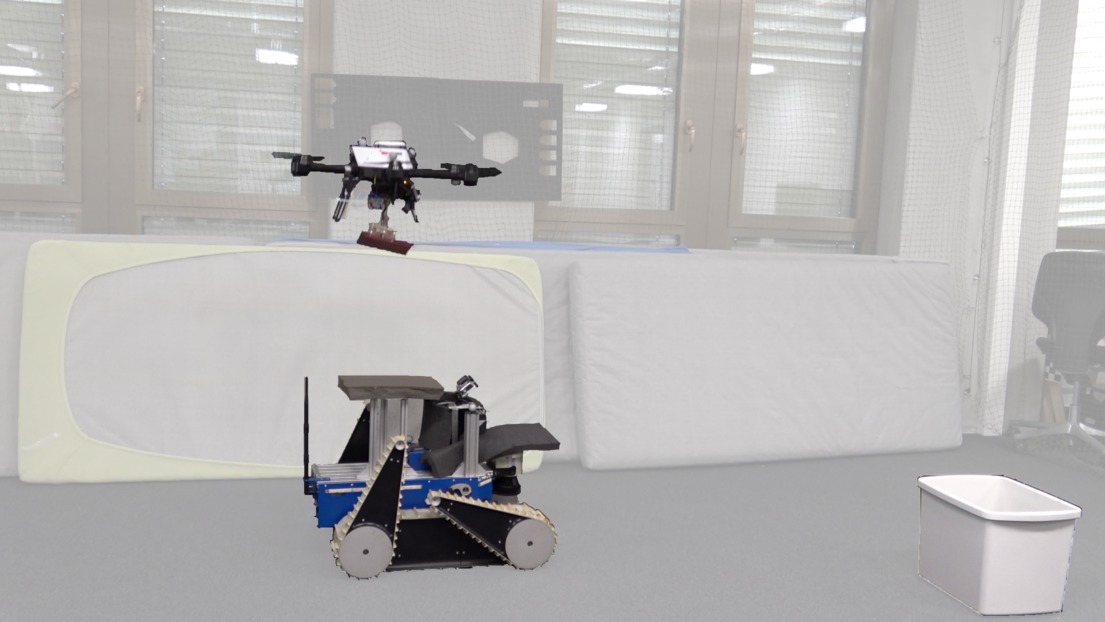}}
\subfloat[\label{fig:teaser_d}]{
\includegraphics[width=0.22\textwidth]{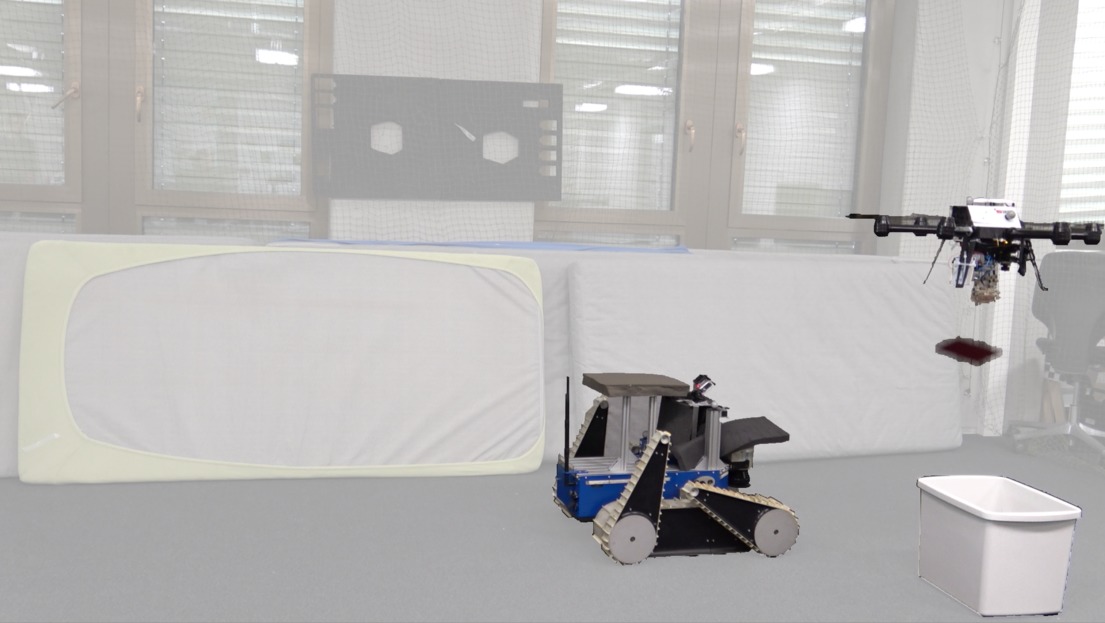}}
\caption{Sequence of the autonomous aerial delivery approach: ~\protect\subref{fig:teaser_a} \ac{MAV} detects object on moving platform and initiates servo positioning. ~\protect\subref{fig:teaser_b} Object is picked using a passively compliant electro-permanent magnetic gripper. ~\protect\subref{fig:teaser_c} \ac{MAV} returns to operation height with object attached and travels to delivery zone. ~\protect\subref{fig:teaser_d} Object is dropped into delivery container by deactivating the electro-permanent magnet after a short hover over the target location.}
\label{fig:teaser}
\end{figure}

A second important challenge for aerial gripping is the correct positioning of the \ac{MAV} towards an object of previously unknown shape and location, and deciding on a successful control for picking such objects. Here, servo-positioning techniques can enable a \ac{MAV} to pick an object by providing relative localization to the object and a controller combined with an approach strategy to yield robust object picking.

We present a novel system that is using electro-permanent magnets and is able to robustly and energy-efficiently pick and deliver stationary or moving objects with a partly ferrous surface of different shapes using a \ac{MAV}.

This work is furthermore motivated by our participation in the \ac{MBZIRC} in which \ac{MAV}s are tasked to autonomously search a field for objects, pick them up and deliver them to a designated drop zone.

Our main contributions are:
\begin{itemize}
\item A low complexity and energy efficient electro-permanent gripper design that allows robust gripping with positional offset and different object shapes.
\item A real-time servo positioning of the \ac{MAV} towards the object.
\item An evaluation of the fully integrated system on different types of objects and in different conditions.
 \end{itemize}


\section{RELATED WORK}
We focus our review of related work on recent advances in aerial gripping and servo positioning techniques for reliably detecting and approaching objects using a \ac{MAV}.

\subsection{Aerial Gripping}

In \cite{ghadiok2011autonomous} the authors propose an integrated object detection and gripping system for \ac{MAV}s using IR diodes for detection and a mechanical gripper for gripping stationary objects. In contrast, our system aims to detect objects using a standard RGB camera and also grip moving objects with an partly ferrous surface.

Transportation of objects using \ac{MAV}s was reported in \cite{michael2011cooperative, maza2010multi, ritz2013carrying}. However, the authors mainly focus on the control of \ac{MAV}s transporting objects. In contrast to our work they do not implement a grip and release mechanism which is an important aspect for fully autonomous delivery.

An aerial manipulation task using a quadrotor with two \ac{DOF} robotic arm was presented in \cite{kim2013aerial}. The kinematic and dynamic models of the combined system were developed and an adaptive controller was designed in order to perform a pick and place task. Such system offers high manipulability, however, the shape of the objects to be picked is limited since the robotic arm is only able to pick thin objects in specific configurations, i.e., thin surfaces pointing upwards. Furthermore, this work assumes that the position of the object to be picked is known in advance.

A self-sealing suction technology for grasping was tested in \cite{kessens2016versatile}. A system capable of grasping multiple objects with various textures, curved and inclined surfaces, was demonstrated. Despite being able to achieve high holding forces, the gripping system requires a heavy compressor and an activation threshold force to pick up the objects. Also, all the tests were performed using a motion capture system with known object positions.

Another type of mechanical gripper was shown in \cite{mellinger2011design}. The gripper uses servo motors to actuate the pins that penetrate the object and create a strong and secure connection. Similar design was also presented in \cite{augugliaro2014aerial}. The main limitation of such gripper is its restriction to pick only objects with penetrable surface. Furthermore, if the surface is not elastically deformable, the gripper might cause irreversible damage to the object.

In \cite{hawkes2016three}, a bio-inspired mechanical gripper was designed in order to allow quadcopters to carry objects with large flat or gently curved surfaces. In addition to being small and light, the gripper consists of groups of tiles coated with a controllable adhesive that allows for very easy attachment and detachment of the object. Nevertheless, the gripper is limited to smooth surfaces, requires tendon mechanism for attachment, and has a limited payload.

OpenGrab EPM\footnote{http://nicadrone.com/} is a gripper developed using the principle of electro-permanent magnets \cite{knaian2010electropermanent}. It is a low-weight, energy efficient and high-payload solution developed for robotic applications and because of its advantages, we have decided to use the same principle for our own gripper. Since OpenGrab EMP is only able to pick flat surfaces, we have developed a more sophisticated design which allows our gripper to pick objects with curved surfaces, while maintaining an equal load distribution on all contacts between object and gripper.

\subsection{Visual Servoing}

\ac{VS} is a well established technique where information extracted from images is used to control the robot motion \cite{pissard1995applying, malis2003robustness, ghadiok2011autonomous}. There are many approaches to deal with \ac{VS}, however some of the most popular include:
\subsubsection{Image Based Visual Servoing}
In this approach, the control law is based entirely on the error in the image plane, no object pose estimation is performed. In \cite{sa2014inspection} the authors employ this method to perform pole inspection with \ac{MAV}s, while in \cite{thomas2016visual} it is used to bring a \ac{MAV} to a perching position, hanging from a pole.

\subsubsection{Pose Based Visual Servoing}
In this approach, the object pose is estimated from the image stream, then the robot is commanded to move towards the object to perform grasping or an inspection task for instance \cite{marchand1999robust}.

Our approach differs from the previous work in the sense that we apply servo positioning for gripping both static and moving object.


\section{Electro-permanent magnetic gripper}

The proposed gripper features two main physical components, i.e., an electro-permanent magnet with electronics board and a passively compliant mechanical structure.

\subsection{Electro-permanent magnet}

The concept of an electro-permanent magnet is based on the physical properties of two different permanent magnets \cite{knaian2010electropermanent}. We consider Alnico and Neodymium magnets. The key properties are their remanence, which is the remaining magnetization after the removal of an external magnetic field and intrinsic coercivity, a measure for the necessary magnetic field to magnetize or demagnetize the material, see Table~\ref{tab:magnets}.

\begin{table}
\begin{center}
\caption{Properties of the magnetic material.}
\label{tab:magnets}
\begin{tabular}{lrr}
\toprule
Material & Remanence & Intrinsic coercivity\\
\midrule
Grade 5 Alnico & $1.25 \unit{T}$ & $48 \unit{kAm^{-1}}$\\
Grade N45 Neodymium & $1.36 \unit{T}$ & $836 \unit{kAm^{-1}}$\\
\bottomrule
\end{tabular}
\end{center}
\end{table}

\begin{figure}
\centering
\subfloat[\label{fig:magnetic_principle}]{
\includegraphics[width=0.2\textwidth]{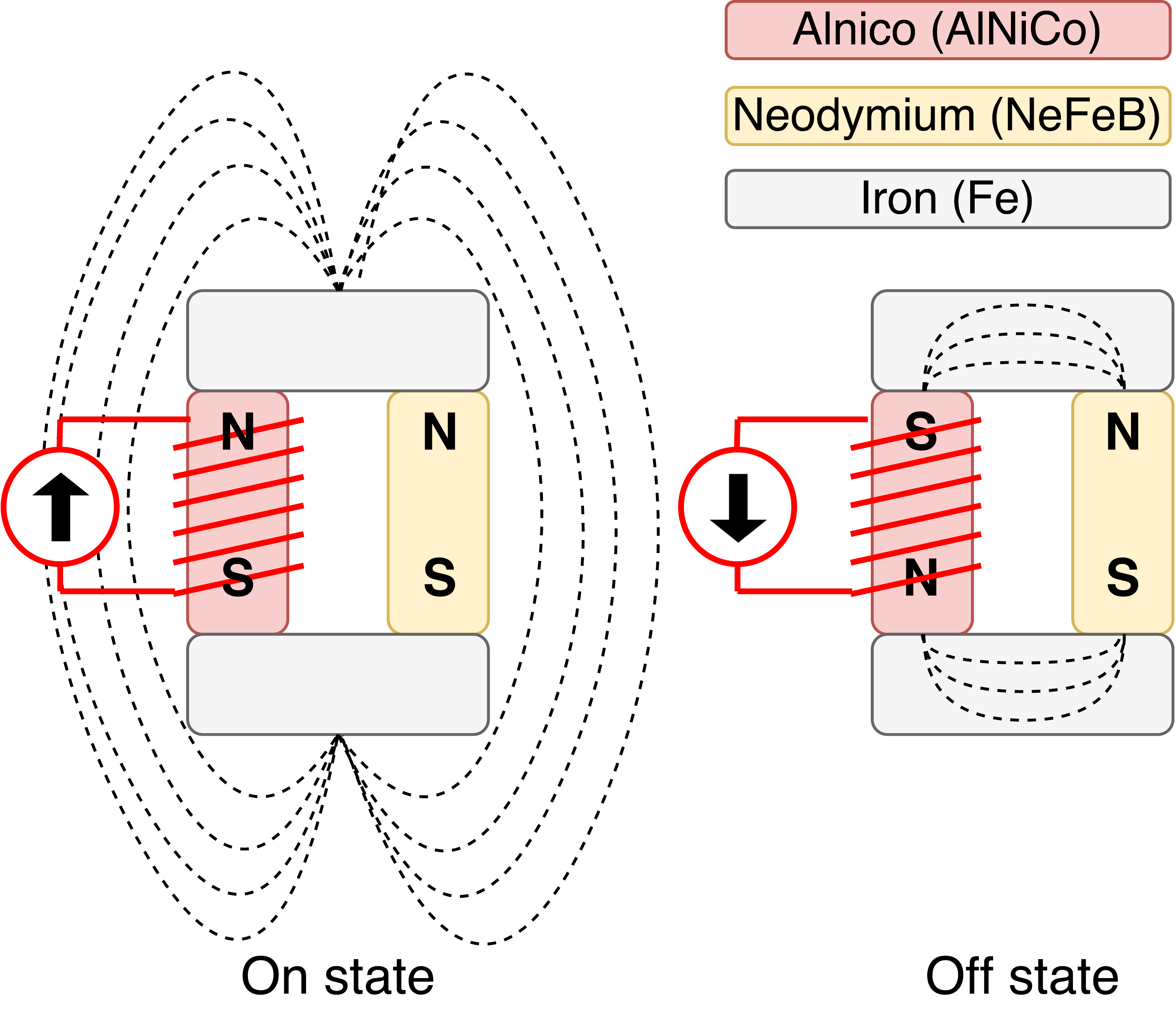}}
\subfloat[\label{fig:magnetic_circuit}]{
\includegraphics[width=0.2\textwidth]{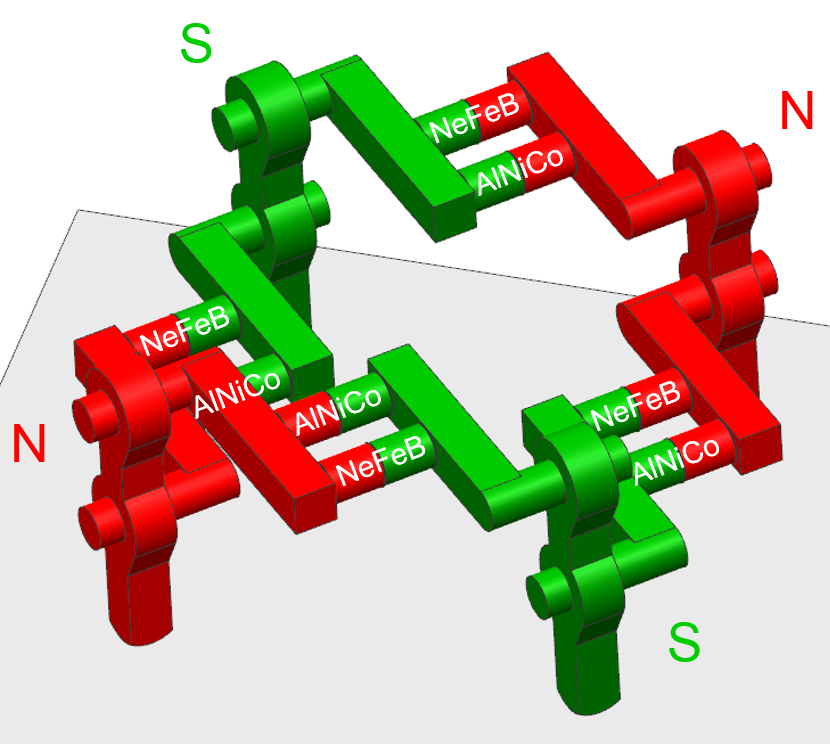}}
\caption{~\protect\subref{fig:magnetic_principle} The figure illustrates the electro-permanent magnet principle. ~\protect\subref{fig:magnetic_circuit} Application of the electro-permanent magnetic principle in two circles as implemented for the design.}
\label{fig:magnet}
\end{figure}

In an electro-permanent magnet, both magnets are assembled in parallel while a coil is winded around the magnet with low intrinsic coercivity, here Alnico. Both magnets are connected to an iron carrier material, as illustrated in Fig.~\ref{fig:magnetic_principle}. Sending a current pulse to the coil generates a magnetic field inside the coil which can switch the magnetic polarization of the Alnico, depending on the direction of the applied magnetic field. The Neodymium magnet stays magnetized in one direction throughout. If both magnets are magnetized in the same direction, the assembly acts as a permanent magnet to the outside, and if they are magnetized in opposite directions the magnetic field is circulating in the assembly and therefore does not act as a magnet to the outside.

\subsection{Mechanical structure}

\begin{figure}
\centering
\subfloat[
\label{fig:assembly}]{\includegraphics[width=0.2\textwidth]{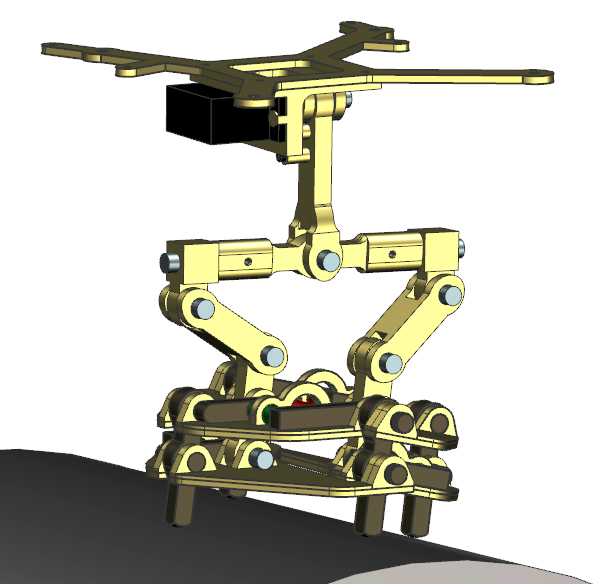}}
\subfloat[\label{fig:parallelogram}]{
\includegraphics[width=0.2\textwidth]{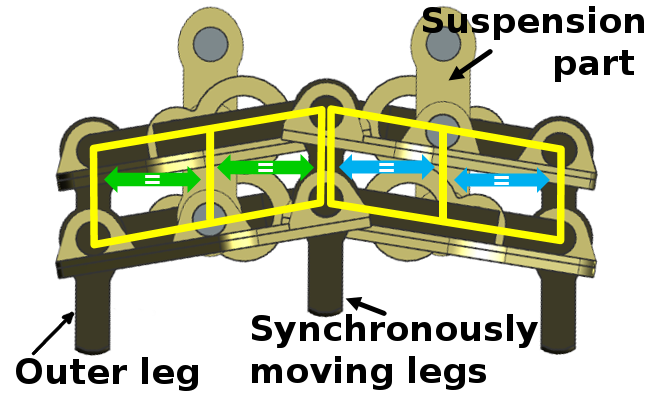}}\\
\subfloat[\label{fig:upper_suspension}]{
\includegraphics[width=0.2\textwidth]{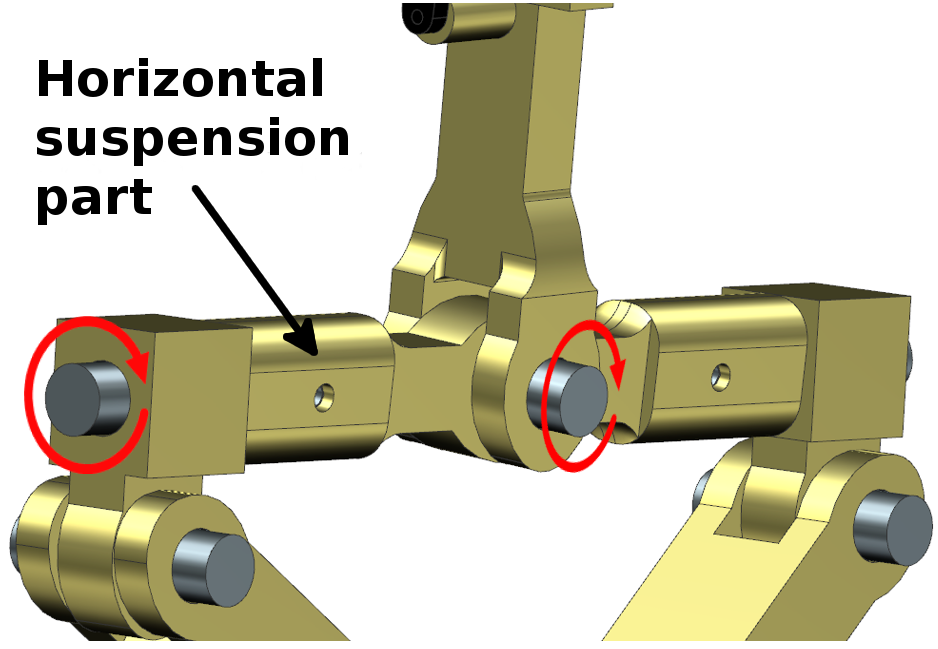}}
\subfloat[\label{fig:retracting}]{
\includegraphics[width=0.2\textwidth]{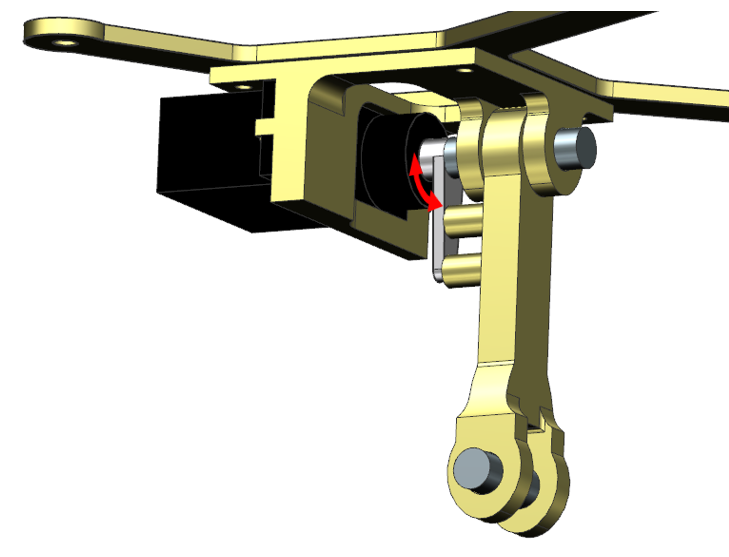}}
\caption{~\protect\subref{fig:assembly} Full assembly of the gripper on a convex surface. ~\protect\subref{fig:parallelogram} Parallelogram structure of end effectors for equal force distribution among all four contacts points between gripper and object. ~\protect\subref{fig:upper_suspension} Upper suspension to provide degrees of freedom in  pitch and roll. ~\protect\subref{fig:retracting} Retracting mechanism for pulling up the  gripper.}
\label{fig:gripper}
\end{figure}

The design of the mechanical structure aims to fulfill three main objectives, i.e., passive adaptivity to different surface geometries, integration of the electro-permanent magnets and functional connectivity to the \ac{MAV}.
We decided to implement 2 cycles of magnetic circuit on the gripper in order to realize a four-point contact to objects, ensuring secure hold. This is illustrated in Fig.~\ref{fig:magnetic_circuit}. For gripping objects of different shapes the functional parts are mounted on a carrier structure that allows for relative motion between the magnetic legs. The full design is illustrated in Fig.~\ref{fig:assembly}. Equal force distribution between the legs is achieved by implementing a parallelogram-shaped support structure, as illustrated in Fig.~\ref{fig:parallelogram}. A suspension with degrees of freedom in pitch and roll enables the gripper to account for attitude changes of the \ac{MAV}, see Fig.~\ref{fig:upper_suspension}. To enable the gripper to extend below the \ac{MAV}'s feet, a retraction mechanism actuated by a servo motor is attached to the upper gripper suspension, see Fig.~\ref{fig:retracting}.

The current design is calculated to a holding force of approximately $34 \unit{N}$ which is tailored to the payload limit and the dynamics of the \ac{MAV} considered. Furthermore, the attraction force can easily be scaled up with a moderate increase in energy consumption and weight of the electro-permanent magnetic components.

\section{Servo Positioning}

The servo positioning module deals with the challenge of autonomously approaching and gripping a detected object. In a first step the \ac{MAV} visually detects an object and localizes its relative pose to the object. Then a controller is activated to yield a desired $x-$, $y-$ position. In the following step the \ac{MAV} executes a strategy for approaching the object in $z-$direction. Finally the \ac{MAV} returns to its operation height and travels to a drop zone, where it releases the object. The drop zone is in a known location.

\subsection{Relative localization}

In our approach we use a simple frame-to-frame detector to estimate the object's \ac{CoG}. Then, we estimate the relative transformation ${}_W\boldsymbol{T}_{m,o}$ between the \ac{MAV} and the object's \ac{CoG} in a global reference frame $W$. For the relative localization step, the object is approximated to have a flat surface for this calculation and the \ac{CoG} to lie in the top plane of the object. The subscripts $m$ and $o$ denote the \ac{MAV} location and the object location respectively.

The \ac{MAV} uses its relative height estimate above the object $h$ and attitude estimate ${}_{W}\boldsymbol{R}_{m}$. The location of the object is then estimated by first calculating the relative rotation ${}_W\boldsymbol{R}_{c,oi}$ between camera center $c$ and object \ac{CoG} in the normalized image plane $oi$ via the vector ${}_{C}\boldsymbol{R}_{c,oi}$ in camera frame $C$. This rotation is transformed into the world coordinate frame using the \ac{MAV}'s attitude estimate  and the rotation between \ac{MAV} base frame and camera center ${}_{M}\boldsymbol{R}_{c}$ in the \ac{MAV} frame $M$, yielding the relative rotation ${}_{W}\boldsymbol{R}_{c,oi}$ in the global coordinate frame $W$.

\begin{equation}
{}_{W}\boldsymbol{R}_{c,oi} = {}_{W}\boldsymbol{R}_{m} \cdot {}_{M}\boldsymbol{R}_{c} \cdot {}_{C}\boldsymbol{R}_{c,oi}
\end{equation}

Using the relative attitude, we calculate the translational component of the offset ${}_{C}\boldsymbol{t}_{oi}$.

\begin{equation}
{}_{C}\boldsymbol{t}_{oi} = -\frac{h}{\begin{pmatrix}0 & 0 & 1\end{pmatrix} {}_{W}\boldsymbol{R}_{c,oi}} \cdot {}_{C}\boldsymbol{R}_{c,oi}
\end{equation}

Finally we calculate the relative translation ${}_{W}\boldsymbol{t}_{m,o}$ between the \ac{MAV} and \ac{CoG} of the object in the global coordinate frame.

\begin{equation}
{}_{W}\boldsymbol{t}_{m,o} = {}_{W}\boldsymbol{R}_{m} ({}_{M}\boldsymbol{t}_{m,c} + {}_{M}\boldsymbol{R}_{c} \cdot {}_{C}\boldsymbol{t}_{oi})
\end{equation}

Here ${}_{M}\boldsymbol{t}_{m,c}$ denotes the calibrated translation between \ac{MAV} base frame and camera center.

The relative transform ${}_W\boldsymbol{T}_{m,o}$ between \ac{MAV} and object is then

\begin{equation}
{}_W\boldsymbol{T}_{m,o} =
\begin{pmatrix}
{}_W\boldsymbol{R}_{m,o} && {}_W\boldsymbol{t}_{m,o} \\
\boldsymbol{0} && 1
\end{pmatrix}
\end{equation}

One of the advantages provided by the design of our gripper is that we can simplify the calculation for planar objects as non-planar surface shapes will be passively handled by the mechanical structure of the gripper.

\subsection{Approaching / Servoing}

The $x,y$-offset and the $z$-offset are handled separately. The $x,y$-offsets are handled by a PID-controller. Based on the error between the object's \ac{CoG} detected in the image frame and the gripper position, a $ \Delta x, \Delta y $ command is generated and added to the current \ac{MAV} pose. The control input is saturated and an anti-windup scheme is implemented. The newly generated \ac{MAV} pose is then tracked by a trajectory tracking \ac{MPC} \cite{mpc2016kamel}. In the case of several objects being present in the \ac{MAV}'s current field of view, the strategy is to first target the object closer in the Euclidean $x,y$-distance.

\begin{figure}
\centering
\includegraphics[width=0.46\textwidth]{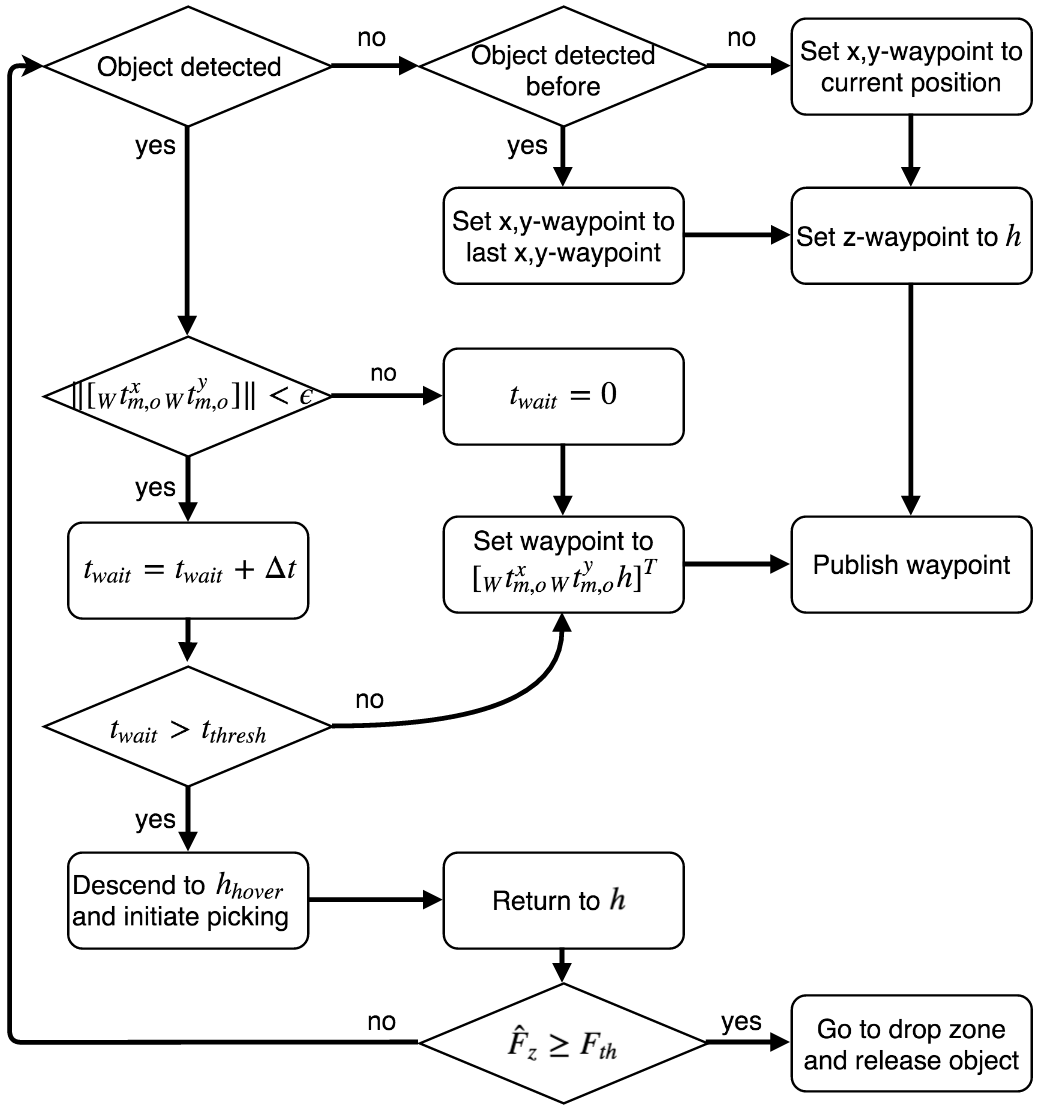}
\caption{State machine of the \ac{MAV} for object picking and delivery.}
\label{fig:state}
\end{figure}

As we assume objects to be on the ground, we define a set approach strategy to yield robust system performance, this is illustrated in Fig.~\ref{fig:state}. The approach-strategy is triggered when the \ac{MAV} stays within a radius $\varepsilon $ in $x,y$ around a point at height $h$ above the \ac{CoG} of the object. If the \ac{MAV} manages to stay within this radius for a set time $t_{wait} > t_{thresh}$, it then descends to a lower height $h_{hover}$ above the object. The servo positioning checks whether the \ac{MAV} is still within a sphere $\varepsilon $ above the object's \ac{CoG} and then initiates the final approach towards the object, which is a guided sequence of descending to the object before ascending to the operation height $h$. If the object is moving during the first descend, the \ac{MAV} uses the previous velocity in $x,y$ for its final descend making it possible to approach linearly moving objects.

If the \ac{MAV} loses sight of the object in the approach sequence, it returns to height $h$ and re-localizes the object in a wider field of view. Please note that a global search strategy is out of the scope of this paper.

\subsection{Delivery}

After successful gripping, the \ac{MAV} flies to the drop zone of known location and releases the object.

An important aspect for robust aerial gripping and transportation is sensing successful gripping of the object. Given that a model-based external disturbances observer based on Extended Kalman Filter (EKF) is employed by the trajectory tracking controller \cite{mpc2016kamel} to compensate for external forces, we decided to employ it to detect successful grasping as well. A successful grasping is detected if the following equation holds:
\begin{equation}\label{eq:gripping_detection}
\hat{F}_{z} \geq F_{th}
\end{equation}
where $ \hat{F}_{z} $ is the $ z $ component of the estimated external force expressed in world frame and $ F_{th} $ is a user defined threshold.


In case the object is lost during transport, the \ac{MAV} returns to the location in which it detected the loss, re-detects the object and performs the servo-positioning from the start. The exact behavior is also shown in Fig.~\ref{fig:state}.



\section{EVALUATION}
The evaluation of our system is three-fold. We test the magnetic behavior of the gripper in simulation and real world experiments, perform a functional evaluation of the gripping with offsets and test the full system under varying conditions, i.e., external disturbances, differently shaped objects and moving objects.

\subsection{Magnetic gripper behavior}

\begin{figure}
\centering
\subfloat[\label{fig:magnet_simulation_off}]{
\includegraphics[width=0.225\textwidth]{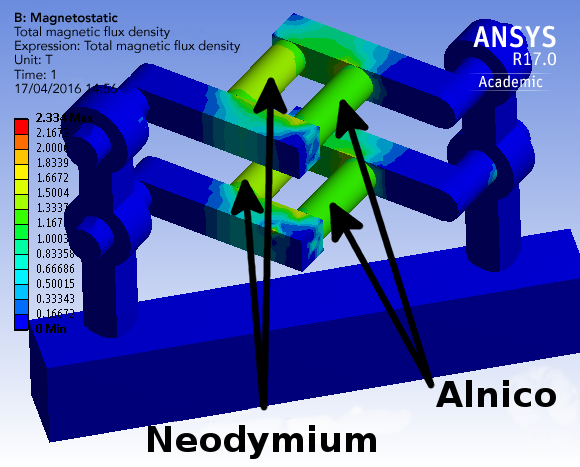}}
\subfloat[\label{fig:magnet_simulation_on}]{
\includegraphics[width=0.225\textwidth]{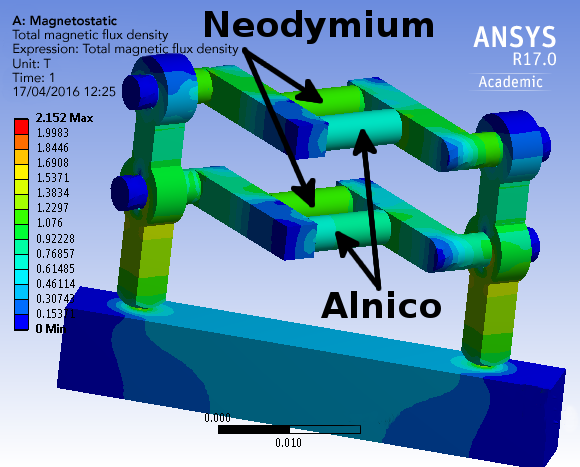}}
\caption{Simulation of the magnetic flux for the electro-permanent magnet, ~\protect\subref{fig:magnet_simulation_off} in the off state forming a closed magnetic circuit inside the material and ~\protect\subref{fig:magnet_simulation_on} in the on state inducing a magnetic field outside the gripper material. The magnetic flux density ranges from $0\unit{T}$ (blue) over $1\unit{T}$ (green) to $2\unit{T}$ (red).}
\label{fig:magnet_simulation}
\end{figure}

Simulation of the magnetic flux is shown in Fig.~\ref{fig:magnet_simulation} for the gripper in the on and off state. With this configuration each of the 2 magnetic cycles of the gripper generates a force of $17\unit{N}$ per magnetic cycle while assuming air gaps between the functional part as depicted in Table~\ref{tab:air_gaps}.

\begin{table}
\begin{center}
\caption{Air gaps in magnetic flux simulation.}
\label{tab:air_gaps}
\begin{tabular}{lr}
\toprule
Loaction & Gap width\\
\midrule
Between magnet and horizontal iron part & $50\unit{\mu m}$\\
Between leg and horizontal iron part & $25\unit{\mu m}$\\
Between leg and object surface & $100\unit{\mu m}$\\
\bottomrule
\end{tabular}
\end{center}
\end{table}

\begin{figure}
\centering
\begin{minipage}[b][.6\linewidth]{0.45\linewidth}
  \centering
  \subfloat[\label{fig:real_assembly}]{%
    \includegraphics[width=\linewidth]{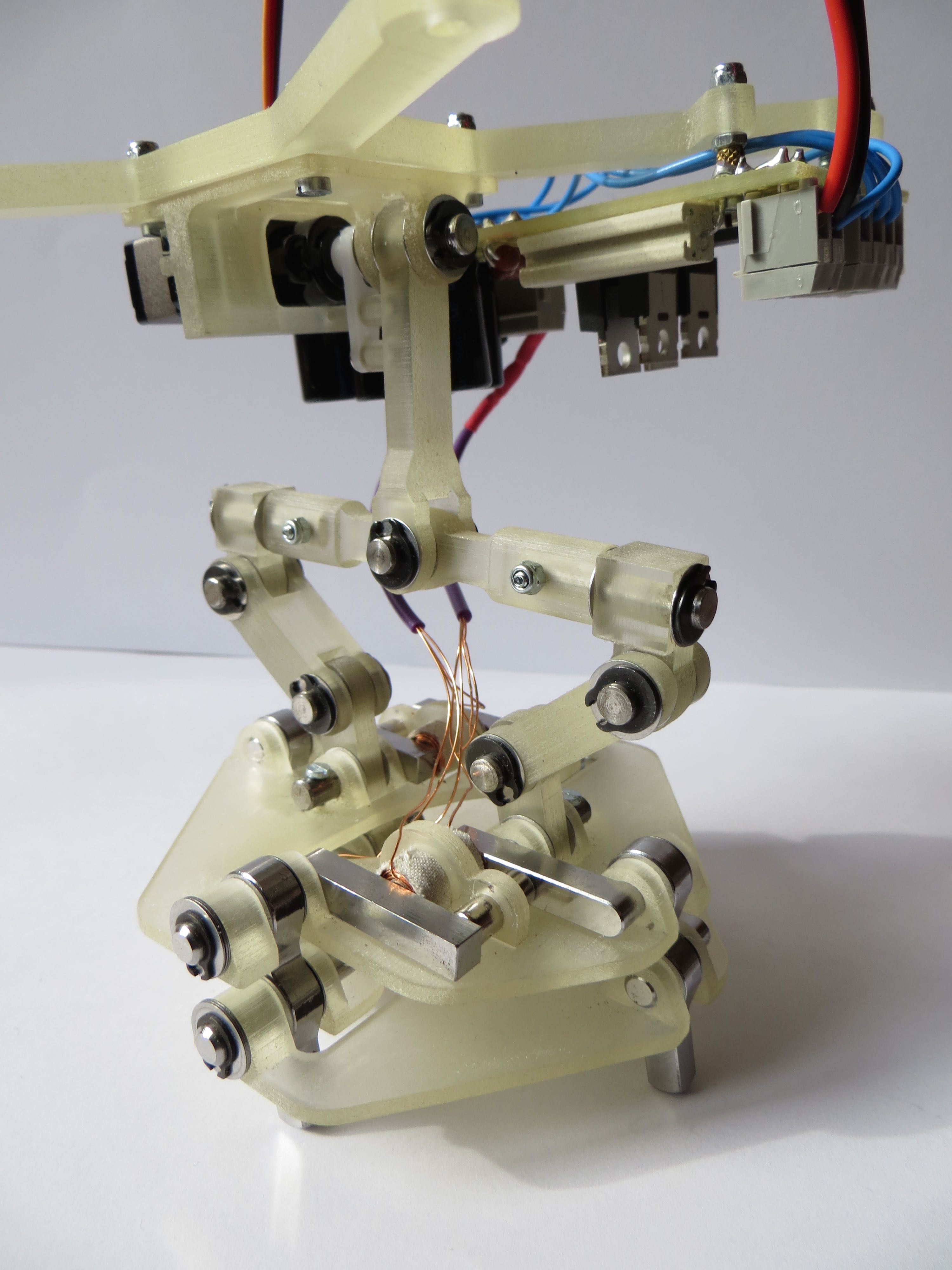}%
  }
\end{minipage}
\begin{minipage}[b][.6\linewidth]{0.5\linewidth}
  \centering
  \subfloat[\label{fig:real_magnets}]{%
    \includegraphics[width=\linewidth]{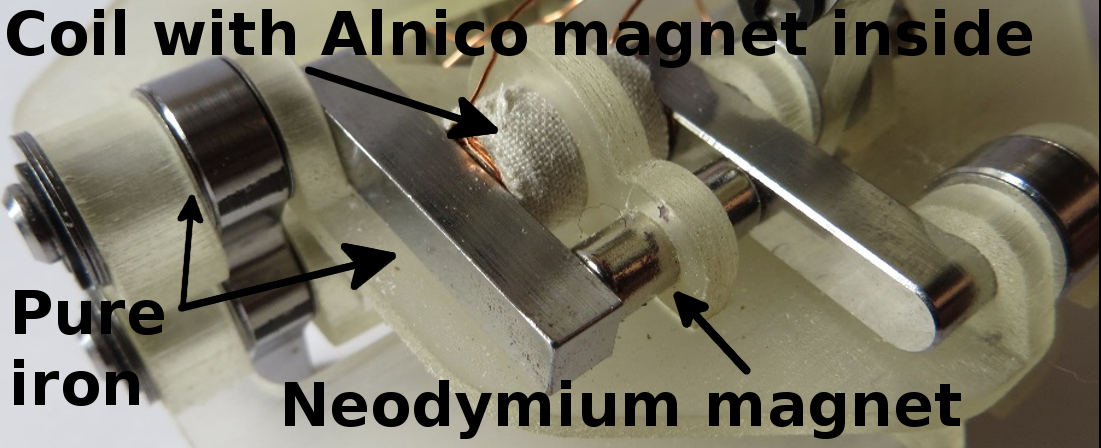}%
  }
  \vfill
  \subfloat[\label{fig:objects}]{%
    \includegraphics[width=\linewidth]{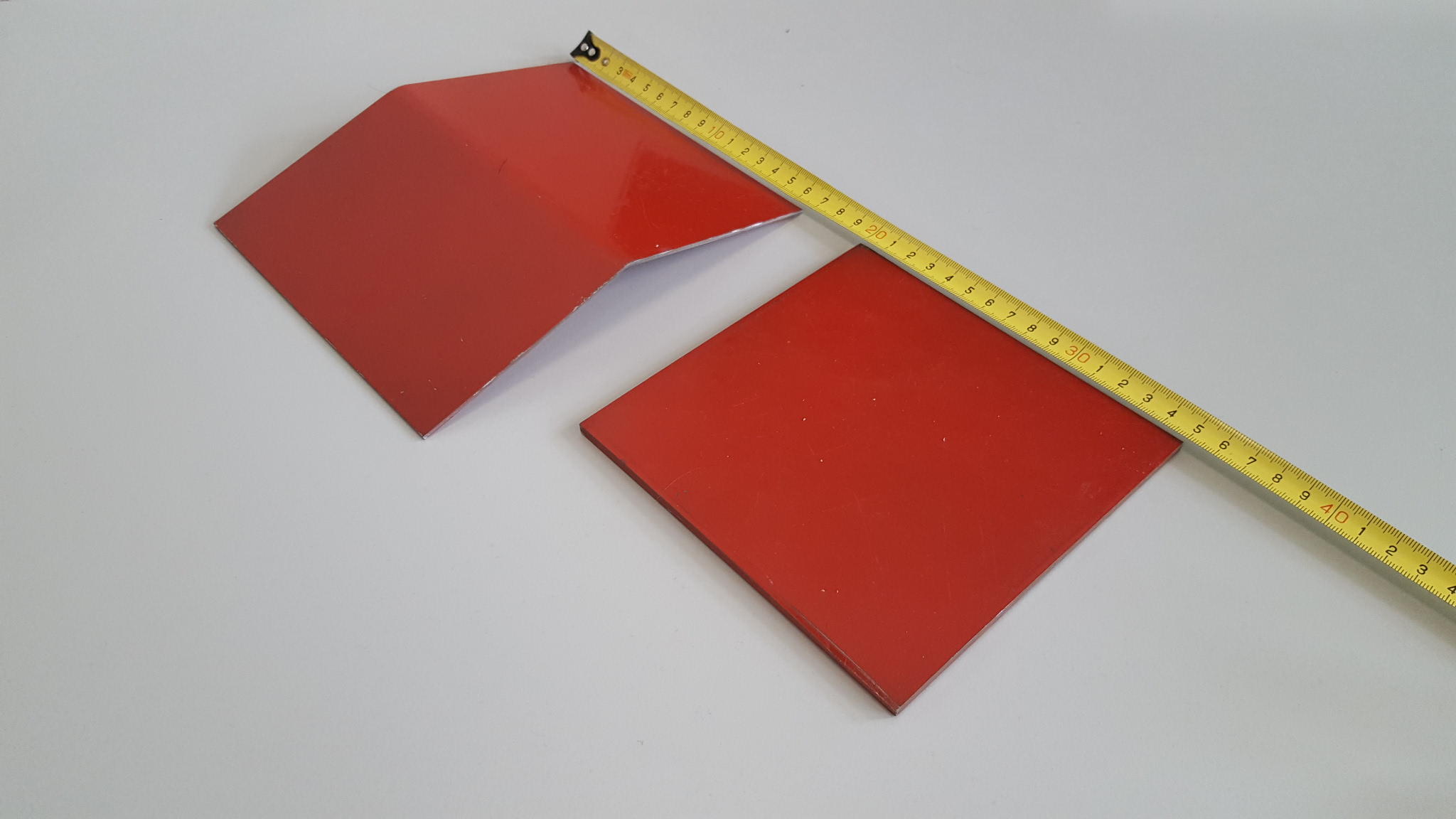}%
  }
\end{minipage}
\caption{Physically realized electro-permanent magnetic gripper: ~\protect\subref{fig:real_assembly} Full assembly of gripper, detached from \ac{MAV}. ~\protect\subref{fig:real_magnets} One of in total four Alnico / Neodymium assemblies in final gripper prototype. ~\protect\subref{fig:objects} Objects used in isolated and integrated tests.}
\label{fig:real_gripper}
\end{figure}

The physical gripper is illustrated in Fig.~\ref{fig:real_assembly}. Tests with the full assembly show that the gripper produces an attraction force of approximately $30\unit{N}$ which is lower than the simulated value. We believe this is due to imperfect manufacturing of the gripper, resulting in slightly different air gaps in the assembly. Nevertheless, the force is still well within acceptable bounds. The functional parts of one of the magnetic cycles is illustrated in Fig.~\ref{fig:real_magnets}. In order to switch between the gripper's on and off state, using the \ac{MAV}'s onboard $15\unit{V}$ batteries, a short $2.5\unit{ms}$ current pulse of $80\unit{A}$ is sent each time, resulting in consumption of $0.8\unit{mWh}$ per switch. The final assembly weighs $210\unit{g}$ including all functional components. However, the materials and the design of the support structure are not optimized yet, especially since we facilitate 3D printed plastic which requires considerably thicker parts to provide the required stiffness compared to light-weight composite materials. Furthermore, a circuit board for fast prototyping was used for the electronic parts, adding $120\unit{g}$ to the weight of the gripper assembly, although developing a \ac{PCB} would significantly decrease this weight.

\subsection{Offset gripping}

We evaluated the gripper assembly in an isolated experiment, using different square metal objects to evaluate its offset gripping behavior. The testing procedure defines a gripping procedure in an offset position in $x,y$ to the objects's \ac{CoG} followed by a vertical acceleration of $0.8g$. The test objects vary in weight and shape, i.e., we test the system on one heavy flat square metal plate and a lighter square metal plate with a bend of $30\degree$ in the middle, as illustrated in Fig.~\ref{fig:objects}. The tests were conducted by linearly increasing the offset of the grip position until the border of the object was reached. The results are illustrated in Table~\ref{tab:offset}.

Although the objects can be lifted statically regardless of the offset position, the vertical acceleration causes both tested objects to show a cutoff offset, i.e., the object is always lost in the lifting when this offset is exceeded. The major causes for losing contact with the object during the dynamic lifting is lateral slipping for the large, but lighter part and loss of contact with the innermost magnetic leg for the smaller, but heavier part. The failure mechanisms can be explained mechanically as follows. The large part starts slipping as the large offset gripping position causes the gripper suspension to have considerable offset rotation in pitch. This causes a lateral force on the magnetic legs that exceeds the friction induced by the magnetic attraction force at the contacts between the legs and the object, causing slippage. The small part fails, because the force is shifted onto the innermost leg due to the leverage effect, causing the combined force of gravity and acceleration to exceed the magnetic attraction force.

\begin{table}
\begin{center}
\caption{Results of offset gripping tests.}
\label{tab:offset}
\begin{tabular}{lrrcrr}
\toprule
\multicolumn{3}{c}{Object} & & \multicolumn{2}{c}{Gripping max.\ values}\\
\cmidrule{1-3}\cmidrule{5-6}
Dimensions & Weight & Bend & & Offset & Pitch angle\\
\midrule
$165 \times 165 \times 2\unit{mm}$ & $520\unit{g}$ & $30\degree$ & & $81\unit{mm}$ & $55\degree$\\
$150 \times 150 \times 4\unit{mm}$ & $870\unit{g}$ & $0\degree$ & & $30\unit{mm}$ & $27\degree$\\
\bottomrule
\end{tabular}
\end{center}
\end{table}

However, the \ac{MAV} controller can reliably provide positional accuracy in $x,y$ which is within the safe bounds of the evaluated gripper's behavior and can therefore provide safe means of gripping in the integrated system's context.
Furthermore, we implemented a gripping detection in the integrated system, causing the \ac{MAV} to re-attempt in case of unsuccessful gripping. Finally, we note that the tested accelerations in this experiment are higher than the ones in the integrated system.

\subsection{Object detection}

For the object detection in the integrated evaluation, we use a simple frame-to-frame recognition scheme using a down-facing camera that is rigidly attached to the base of the \ac{MAV}. Objects are assumed to be arbitrarily shaped and of red, blue or black color. We aim to detect the \ac{CoG} of the objects. Therefore, the images are undistorted, down-sampled to $\frac{1}{8}$ resolution and converted to HSV color space.
The detector performs morphological opening to remove small foreground detections, and morphological closing to fill small holes in the foreground. Then it detects contours on the binary images, which are filtered by a threshold to reject very small (contour including $ <0.4\% $ of the image plane) and very large objects (contour including $ >90\% $ of the image plane). The remaining contours are detected as objects and their \ac{CoG} calculated by using their 0th and 1st moments. The thresholding is performed on all three axes of the HSV color space to distinguish between red, blue and black objects.

\subsection{Integrated System Evaluation}

\begin{figure}
\centering
\includegraphics[width=0.46\textwidth]{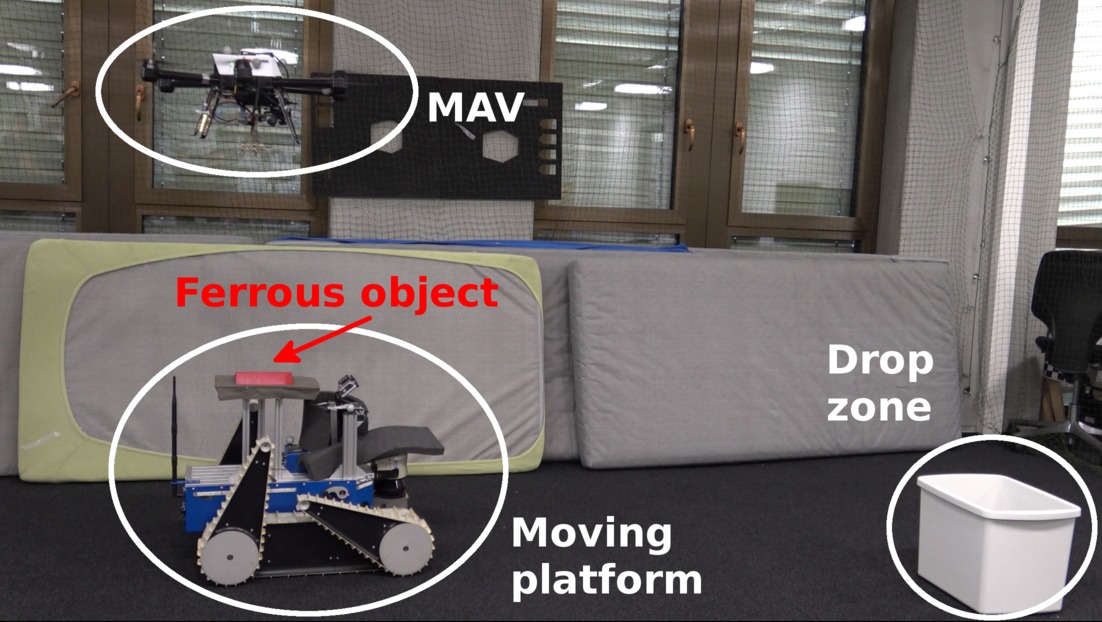}
\caption{Experimental setup for the integrated evaluation, here for the dynamic experiment. The \ac{MAV} is tasked to autonomously detect and grip the ferrous object that is placed on the moving platform. After successfull grasping, the object is delivered to the drop zone.}
\label{fig:scenario}
\end{figure}

A \ac{MAV} equipped with a down-facing Pointgrey Chameleon3 $3.2\unit{MP}$ camera with a fisheye lens running at $20\unit{Hz}$ and the presented gripper were evaluated in an indoor motion capture room, as illustrated in Fig.~\ref{fig:scenario} and video supplement\footnote{https://goo.gl/6uZH0M}. In a first experiment a bend ferrous object is placed in a random location across the room in one of two configurations, i.e., either with the bend facing upwards or downwards, providing a convex or concave object to pick. Then the \ac{MAV} is brought manually to a hover in a random location with the object in its field of view. The \ac{MAV} is then tasked to autonomously execute the object detection, servo positioning, gripping and transportation to a known drop-off location for releasing the object. This experiment is repeated 23 times using the bend metal object in different configurations. We also perform 5 trials while applying varying strengths of wind to the platform of up to $15\unit{m/s}$. Furthermore, another experiment is executed 14 times with the object placed on a moving platform that moves linearly in an arbitrary direction with a set velocity of $0.1\unit{m/s}$ which was not communicated to the \ac{MAV}. We use the AscTec Neo \ac{MAV} for these experiments and a motion capture system for tracking the \ac{MAV}. However, the \ac{MAV} is not limited to operate with a tracking system, and alternatively a state estimation on board the \ac{MAV} can be used \cite{weiss2013monocular}. The results of these tests are presented in Table~\ref{tab:integrated_test}. Here we report the success rates along with the number of experiments and the number of pick up trials, i.e., the number of total pick up repetitions if the part is detected to not be picked and a re-picking is triggered. The procedure is illustrated in Fig.~\ref{fig:teaser} for the dynamic experiments. The static experiments were performed in a similar setup with the object being placed on the ground.

\begin{table}
\begin{center}
\caption{Results of integrated system (\ac{IS}) tests.}
\label{tab:integrated_test}
\begin{tabular}{lrrr}
\toprule
Experiment type & Success rate & Experiments & Pick up tries\\
\midrule
\ac{IS}                   & $95.65\,\%$   & $23$ & $25$\\
\ac{IS} + wind            & $100\,\%$     & $5$ & $8$\\
\ac{IS} + dynamic objects & $78.57\,\%$   & $14$ & $27$\\
\bottomrule
\end{tabular}
\end{center}
\end{table}

\subsection{Findings}

Throughout all static experiments we recorded only one failure in the delivery action due to releasing the object next to the drop zone and therefore missing the drop zone container. Since the object landed too close to the container the MAV was not able to pick it up again because of the confined space and risk of crashing with the container. As expected, the picking quality decreased in the case of external disturbances since positioning accuracy achievable by the MAV decreases. Furthermore, in some cases, we noticed that reflections on the object can cause the detector to estimate a \ac{CoG} that is off-centered, thus decreasing positioning accuracy, i.e., two repeated tries. In such cases, the MAV was not able to properly grip the object in the first approach, however, it was still able to detect this, recover, and retry the procedure.

In the dynamic case, when the object is placed on a moving platform we recorded decreased success rate for picking. We counted failure cases if the object was not accurately picked by the \ac{MAV} causing it to slip and fall to the ground. Although the object could be recovered from the ground as static object, we cancelled the experiment in these cases and recorded failure. We noted that, since we perform frame-wise detection, we do not have an accurate velocity estimate of the moving platform which could be improved by implementing an object tracking over time.


\section{CONCLUSIONS \& FUTURE WORK}

In this paper, we have presented a full system for  energy-efficient, autonomous picking and delivery of ferrous objects with \ac{MAV}s. The integrated system is based on gripping technology with electro-permanent magnets.

We have evaluated the core innovations of our pipeline separately and the integrated system as a whole. Our results show that even under varying conditions the \ac{MAV} is able to pick and deliver the objects in the static case and most of the times in the dynamic case as well. In contrast to state-of-the-art approaches which rely either on known object locations, known object shapes or high position accuracy of the \ac{MAV}, our approach can handle all of these unknowns in an integrated manner while achieving very high delivery success rates. Furthermore, the proposed gripper design for \ac{MAV}s combining passive compliance with electro-permanent magnets, to our best knowledge, has not been shown before.

For future work, we plan to further optimize our gripper design towards weight and compliance and integrate the camera in a next version of the gripper as its field of view is partly occluded in the current setup. We furthermore plan to implement object tracking and feed-forward control to increase the system performance for picking of moving objects. Another interesting avenue is a combination of our system with global search strategies and multiple \ac{MAV}s.


\addtolength{\textheight}{-12cm}   




\section*{ACKNOWLEDGMENT}

This work was supported by the European Union's Seventh Framework Programme for research, technological development and demonstration under the TRADR project No. FP7-ICT-609763, European Union’s Horizon 2020 Research and Innovation Programme under the Grant Agreement No.644128, AEROWORKS, and Mohamed Bin Zayed International Robotics Challenge 2017.

\bibliographystyle{IEEEtran}

\bibliography{eth-bib}

\begin{acronym}
\acro{TRADR}{``Long-Term Human-Robot Teaming for Robots Assisted Disaster Response''}
\acro{MAV}{Micro Air Vehicle}
\acro{MBZIRC}{Mohamed Bin Zayed International Robotics Challenge}
\acro{CoG}{Center of Gravity}
\acro{MPC}{Model Predictive Controller}
\acro{IS}{integrated system}
\acro{PCB}{Printed Circuit Board}
\acro{DOF}{degrees of freedom}
\acro{IBVS}{Image based Visual Servoing}
\acro{VS}{Visual Servoing}
\end{acronym}

\end{document}